\documentclass[10pt,conference]{IEEEtran}
\IEEEoverridecommandlockouts
\usepackage{cite}
\usepackage{amsmath,amssymb,amsfonts}
\usepackage{algorithmic}
\usepackage{graphicx}
\usepackage{textcomp}
\usepackage{subcaption}
\usepackage{tikz}
\usepackage{fontawesome}
\usetikzlibrary{shapes}
\usepackage{array}
\newcolumntype{P}[1]{>{\centering\arraybackslash}p{#1}}
\usepackage[htt]{hyphenat}
\usepackage[hyphens]{url}
\usepackage[nolist]{acronym}
\usepackage{xcolor}
\def\BibTeX{{\rm B\kern-.05em{\sc i\kern-.025em b}\kern-.08em
    T\kern-.1667em\lower.7ex\hbox{E}\kern-.125emX}}
\begin{document}

\title{DeepEdgeBench: Benchmarking Deep Neural Networks on Edge Devices*\\
{\footnotesize \textsuperscript{*}Note: This the preprint version of the accepted paper at IC2E'21}
}

\author{Stephan Patrick Baller\IEEEauthorrefmark{1}, \IEEEauthorblockN{Anshul Jindal\IEEEauthorrefmark{1}, Mohak Chadha\IEEEauthorrefmark{1}, Michael Gerndt\IEEEauthorrefmark{1} \\}
\IEEEauthorblockA{\IEEEauthorrefmark{1}Chair of Computer Architecture and Parallel Systems, Technische Universit{\"a}t M{\"u}nchen, Germany \\ Garching (near Munich), Germany \\
Email: \{stephan.baller, anshul.jindal, mohak.chadha\}@tum.de, gerndt@in.tum.de \\}
}

\maketitle
\begin{acronym}[AI]
\acro{SoC}[SoC]{System on a chip}
\acroplural{SoCs}[SoC]{Systems on a chip}
\acro{AI}[AI]{Artificial Intelligence}
\acro{ML}[ML]{Machine Learning}
\acro{DL}[DL]{Deep Learning}
\acro{NN}[NN]{Neural Network}
\acro{DNN}[DNN]{Deep Neural Network}
\acro{CCTV}[CCTV]{Closed-circuit television}
\acro{TF}[TF]{Tensorflow}
\acro{OS}[OS]{Operating System}
\acro{IoT}[IoT]{Internet of Things}
\acro{RAM}[RAM]{Random Access Memory}
\acro{TOPS}[TOPS]{Tera Operations Per Second}
\acro{GFLOPS}[GFLOPS]{Giga Floating Operations Per Second}
\acro{NTP}[NTP]{Network Time Protocol}
\acro{SFTP}[SFTP]{Secure File Transfer Protocol}
\acro{CNN}[CNN]{Convolutional Neural Network}
\acro{CPU}[CPU]{Central Processing Unit}
\acro{GPU}[GPU]{Graphics Processing Unit}
\acro{NPU}[NPU]{Neural Processing Unit}
\acro{TPU}[TPU]{Tensor Processing Unit}
\acro{ASIC}[ASIC]{Application-specific integrated circuit}
\acro{PCIe}[PCIe]{Peripheral Component Interconnect express}
\acro{CUDA}[CUDA]{Compute Unified Device Architecture}
\acro{CUDNN}[CUDNN]{CUDA Deep Neural Network Library}
\acro{MVE}[MVE]{M-Profile Vector Extension}
\acro{API}[API]{Application Programming Interface}
\acro{LAN}[LAN]{Local Area Network}
\acro{FPS}[FPS]{Frames per Second}
\end{acronym}

\begin{abstract}

EdgeAI (Edge computing based Artificial Intelligence) has been most actively researched for the last few years to handle variety of massively distributed AI applications to meet up the strict latency requirements. Meanwhile, many companies have released edge devices with smaller form factors (low power consumption and limited resources) like the popular Raspberry Pi and Nvidia's Jetson Nano for acting as compute nodes at the edge computing environments. Although the edge devices are limited in terms of computing power and hardware resources, they are powered by accelerators to enhance their performance behavior. Therefore, it is interesting to see how AI-based Deep Neural Networks perform on such devices with limited resources. 

In this work,  we present and compare the performance in terms of inference time and power consumption of the four \acsp{SoC}:  Asus Tinker Edge R, Raspberry Pi 4, Google Coral Dev Board, Nvidia Jetson Nano,   and one microcontroller: Arduino Nano 33 BLE, on different deep learning models and frameworks. We also provide a method for measuring power consumption, inference time and accuracy for the devices, which can be easily extended to other devices. Our results showcase that, for Tensorflow based quantized model, the Google Coral Dev Board delivers the best performance, both for inference time and power consumption. For a low fraction of inference computation time, i.e. less than \textit{29.3}\% of the time for MobileNetV2, the Jetson Nano performs faster than the other devices.

\end{abstract}


\section{Introduction}

Advances in computing devices and high-speed mobile networking provide today’s applications to be distributed globally. Such applications are thus deployed on cloud service platforms to benefit from on demand provisioning, unlimited resource pooling, and dynamic scalability. Modern deep learning techniques serve a key component in various real-time applications, like speech recognition~\cite{6296526}, recommendation systems~\cite{10.1145/2736277.2741667} and video classification~\cite{10.1109/CVPR.2014.223}. However, deep learning-based approaches require a large volume of high-quality data to train and are very expensive in terms of computation, memory and power consumption~\cite{zhang2020deep}. Moreover, existing
cloud computing is unable to manage these massively distributed applications and analyze their data due to: i) challenges posed on the network capacity when tasks are deployed to the cloud~\cite{6566921}; ii) many applications, for example, autonomous driving~\cite{8932460}, have strict latency requirements that the cloud would have difficulty meeting since it may be far away from the users~\cite{10.1145/2967360.2967369}.

The concept of Edge Computing has been recently proposed
to complement cloud computing to resolve these problems by performing certain tasks at the edge of the network~\cite{7488250}. The idea is to distribute parts of processing and communication to the "edge" of the network, i.e closer to the location where it is needed. As a result, the server needs less computing resources, the network is less strained and latencies is decreased. 

Edge Devices can come in a variety of forms ranging from large servers to low-powered \ac{SoC} devices like the popular Raspberry Pi or any other ARM based devices. \acp{DNN} may occupy big amounts of storage and computing resources~\cite{OpenAIPresentsGPT32020,TensorflowModels2021}. Although the edge devices are limited in terms of computing power and hardware resources, they are powered by accelerators to enhance their
performance at the edge computing environments.

In the context of Edge Computing, it is rather interesting to see how devices with smaller form factors (low power consumption and limited resources) can handle \ac{DNN} evaluation.  There are a considerable number of articles on the benchmark of edge devices~\cite{10.1145/3363347.3363363, varghese2020survey, 8506339, 9070647, fdn}. However, some of the articles are already outdated and others miss benchmarks on the major latest edge devices, thus lacking a thorough comparison between the edge devices concerning the \ac{DNN} applications. Unlike other works, this article focuses on evaluating the performance of recent edge device for DNN models inferencing. 
The key contributions of this work are as follows:
\begin{itemize}
    \item We present a comparative analysis on recent accelerator-based edge devices specialized for the \ac{DNN} application domain. We choose following devices as our target edge devices to assess their performance behavior and capabilities for \ac{DNN} applications:
\begin{itemize}
	\item ASUS Tinker Edge R~\cite{TinkerEdge}
	\item Raspberry Pi 4~\cite{foundationRaspberryPiModel}
	\item Google Coral Dev Board~\cite{DevBoard}
	\item NVIDIA Jetson Nano~\cite{JetsonNanoDeveloper2019}
	\item Arduino Nano 33 BLE~\cite{ArduinoNano33}
\end{itemize}
    \item Our evaluation perspectives include inference speed for a fixed number of images, power consumption during (accelerated) \ac{DNN} inference, and models accuracies with and without optimization for the respective device.
    \item We used four different deep learning inference frameworks for the evaluation: Tensorflow, TensorRT, Tensorflow Lite and RKNN-Toolkit.
    \item  We also provide a method for measuring power consumption, inference time and accuracy for the devices, which can be easily extended to other devices. We open source the collected data and the developed method for further research\footnote{\url{https://github.com/stephanballer/deepedgebench}}.
    
\end{itemize}

The rest of this paper is organized as follows. \S\ref{sec:target_edge} gives an overview of the target edge devices, model frameworks and formats used in this work for the evaluation. In \S\ref{sec:methodology}, the overall evaluation setup and methodology are described. \S\ref{sec:exp_config} provides the experimental configurations details. In \S\ref{sec:results}, our performance evaluation results are presented along with the discussion of those results. Finally, \S\ref{sec:conclusion} concludes the paper and presents a future outlook.

\section{Target Edge Devices, Frameworks and Model Formats}
\label{sec:target_edge}
In this section, we compare different edge device architectures and present their hardware overview, and also briefly describe the characteristics of different \ac{DNN} models and frameworks used in this work for benchmarking the edge devices.

\subsection{Target Edge Devices}
We have considered the following five edge devices summarized in Table~\ref{tab:devspec}.
\begin{table*}[t]%
\centering
\caption{Overview on different target edge devices specifications}%
\begin{tabular}{ |p{.1\textwidth}|*{5}{p{.145\textwidth}|} }
    \hline
    & \textbf{Tinker Edge R (2019)}~\cite{TinkerEdge} & \textbf{Raspberry Pi 4 (2019)}~\cite{foundationRaspberryPiModel} & \textbf{Google Coral Dev Board (2020)}~\cite{DevBoard} & \textbf{NVIDIA Jetson Nano (2019)}~\cite{JetsonNanoDeveloper2019} & \textbf{Arduino Nano 33 BLE (2019)}~\cite{ArduinoNano33} \\
    \hline
    \acs{CPU} & RK3399Pro, Dual-core Cortex A72 @1.8GHz, Quad-core Cortex A53 @1.4GHz\newline (ARMv8A) & BCM2711, Quad-core Cortex A72 @1.5GHz\newline (ARMv8A) & NXP i.MX 8M,\newline Quad-core Cortex A53 @1.5GHz\newline (ARMv8A) & Cortex A57 @1.43GHz\newline (ARMv8A) & nRF52840, Cortex M4 @64MHz\newline (ARMv7E-M) \\
    \hline
    \ac{AI} Unit & Rockchip NPU & - & Google Edge TPU & 128-core Maxwell GPU & - \\
    \hline
    Memory & LPDDR4 4GB, LPDDR3 2GB (NPU) & LPDDR4 4GB @3200MHz & LPDDR4 1GB @1600MHz & LPDDR4 4GB @1600MHz & 256KB SRAM (nRF52840) \\
    \hline
    Storage & 16GB eMMC, Micro SD & Micro SD & 8GB eMMC, Micro SD & Micro SD & 1MB Flash Memory \\
    \hline
    First Party OSs & TinkerOS, Android & Raspbian & Mendel Linux & Ubuntu & MBed  OS \\
    \hline
    Optimized Frameworks & TensorFlow 1, Tensorflow Lite, Caffe, Kaldi, MXNet, ONNX  & - & TensorFlow Lite & TensorFlow, TensorRT, NVCaffe, Kaldi, MXNet, DIGITS, PyTorch & TensorFlow Lite Micro \\
    \hline
\end{tabular}
\label{tab:devspec}
\end{table*}
\subsubsection{Nvidia Jetson Nano}
The Jetson Nano is one of the offerings from Nvidia for edge computing~\cite{JetsonNanoDeveloper2019}. With the Jetson product line, Nvidia deploys their \ac{GPU} modules to the Edge with accelerated \ac{AI} performance.
The Jetson Nano’s GPU is based on the Maxwell microarchitecture (GM20B) and comes with one streaming multiprocessor (SM) with 128 CUDA cores, which allows to run multiple neural networks in parallel~\cite{JetsonNanoDeveloper2019}. The Jetson Nano is the lowest-end version of the product line with a peak performance of 472 \ac{GFLOPS}~\cite{JetsonNanoBrings2019}. It comes with either 2GB or 4GB \ac{RAM} version wherefore we're testing the 4GB version. The Nano can be operated in two power modes, 5W and 10W. We used in our experiment an Nvidia Jetson Nano with 4 GB RAM and 10 W to maximize performance. Compared to the other target edge devices in this work, the Jetson Nano stands out with a fully utilizable \ac{GPU}.

\subsubsection{Google Coral Dev Board}
Coral is a platform by Google for building AI applications on edge devices~\cite{Coral}. The Google Coral Dev Board is one of the offerings which features the “edge” version of the TPU (tensor processing unit)~\cite{10.1145/3079856.3080246}, an application specific integrated circuit (ASIC) designed for accelerating neural network machine learning, particularly using the TensorFlow framework~\cite{tensorflow2015-whitepaper}. The Edge TPU
operates as a co-processor in an edge device and allows for an efficient aggregation of tens of thousands of ALUs (arithmetic logic units) and faster data transfer rate between the TPU and the memory. However, the Edge TPU is fine-tuned for matrix operations which are very frequent in neural network machine learning.

The Google Coral Dev Board comes with either 1GB or 4GB of \ac{RAM}.
The \ac{SoC} integrates the Google Edge TPU with the performance of 4 trillion operations (tera-operations) per second (TOPS) or 2 \ac{TOPS} per watt~\cite{EdgeTPUPerformance}. To make use of the dedicated unit, models in a supported format can be converted to work with the PyCoral and Tensorflow Lite framework~\cite{TensorFlowModelsEdge}.

\subsubsection{Asus Tinker Edge R}
Currently, ASUS offers six devices under the Tinker Board brand~\cite{TinkerBoards}. Two of them, the Tinker Edge R~\cite{TinkerEdge} and Tinker Edge T~\cite{TinkerEdgeT} are specifically made for \ac{AI} applications. While the Tinker Edge T is supported by a Google Edge TPU, the  Tinker Edge R uses Rockchip \ac{NPU} (RK3399Pro), a Machine Learning (ML) accelerator that speeds up processing efficiency, and lowers power demands. With this integrated Machine Learning (ML) accelerator, the Tinker Edge R is capable of performing 3 tera-operations per second (\ac{TOPS}), using low power consumption. And it’s optimized for Neural Network (NN) architecture, which means Tinker Edge R can support multiple Machine Learning (ML) frameworks and common Machine Learning (ML) models can be easily compiled and run on the Tinker Edge R. Rockchip, the manufacturer of the chip, provides the RKNN-Toolkit as a library to convert and run models from different frameworks. As operating system, Asus offers a Debian Stretch and Android 9 variant of its so called Tinker \ac{OS} to power the board. Compared to the other devices, the board offers additional interfaces for expanding connectivity (mPCIe, M.2, serial camera, serial display), especially interesting for IoT applications.

\subsubsection{Raspberry Pi 4}
The devices from the Raspberry Pi series are among the most popular SoCs and represent go-to products for people who are interested in IoT~\cite{RasphberryProducts}. In June 2019, the Raspberry Pi 4 was released, featuring a variety of \ac{RAM} options and a better processor. The Raspberry Pi 4 comes with Gigabit Ethernet, along with onboard wireless networking and Bluetooth. The Rasbperry Pi is commonly used with Raspbian, a 32 Bit \ac{OS} based on Debian. As the Raspberry Pi 4 is capable of running a 64 Bit \ac{OS}. We're instead going to use the available 64 Bit Raspbian version (aarch64 architecture) to provide the same testing conditions as the other devices. This actually makes a difference in performance, as a blog post~\cite{croceWhyYouShould2020} on benchmarking the Raspberry Pi from 2020 suggests. In contrary to it's predecessors, the 4th generation is available in multiple variants with different \ac{RAM} sizes (2, 4 and 8GB). For this work we're testing the 4GB variant.

While all other devices in this work (excluding the Arduino) include \acp{GPU} or co-processors for neural computation, the Raspberry Pi lacks such a dedicated unit, but stands out with its low price and availability compared to the other products.


\subsubsection{Arduino Nano 33 BLE}
Besides the main A-Profile architecture family from Arm Ltd. company used in \acp{SoC}, the M-Profile for microcontrollers is meant to provide ''low-latency, highly deterministic operation for deeply embedded systems''~\cite{ltdMProfileArchitectures}. In their latest revision Armv8.1-M, a new \ac{MVE} is included to accelerate \ac{ML} algorithms. The Cortex Armv7-M is the latest architecture used in currently available microcontrollers like Arduinos. Additionally, \textit{TinyML} is a foundation aiming to bring \ac{ML} inference to ultra-low-power devices, i.e. microcontrollers~\cite{banburyBenchmarkingTinyMLSystems2021}. Inference frameworks like Tensorflow Lite already implement this approach by enabling \ac{DL} models to be inferenced on products like the Arduino Nano 33 BLE or ESP-32~\cite{TensorFlowLiteMicrocontrollers}.

Therefore, the Arduino Nano 33 BLE is also considered as target edge device to represent this category of edge devices. It features the nRF52840, a 32-bit ARM Cortex™-M4 CPU running at 64 MHz from Nordic Semiconductors~\cite{ArduinoNano33}. It is officially supported by the Tensorflow Lite Micro \ac{DL} framework and counts to the ''larger'' microcontroller systems~\cite{TensorFlowLiteMicrocontrollers}. With a Flash Memory size of 1MB, it's therefore capable of running networks with up to 500KB in size~\cite{ArduinoNano33}.

\subsection{Inference frameworks and conversion between them }
All devices share \ac{TF} as common framework to provide optimized model inference, though only being compatible with certain models and Tensorflow variants. The \ac{TPU} of the Google Coral Dev Board is only available for use with optimized Tensorflow Lite models, RKNN-Toolkit for Tinker Edge R only supports conversion of models in Frozen Graph format and Tensorflow Lite models and TensorRT for the Nvidia Jetson Nano doesn't support Tensorflow Lite. Besides, there are further restrictions for converting models from Tensorflow to be optimized for the respective device.

To make use of the devices' \acp{NPU}, \acp{TPU} and \acp{GPU}, dedicated Python \acp{API}s are provided to convert and inference models from one or more frameworks. In this subsection we describe four such methods. 

\begin{table*}[t]%
\centering
\caption{Optimization compatibility for pretrained Tensorflow models on various edge devices~\cite{RockchipQuickStart,TensorRTDeveloperGuidePdf}}
    \begin{tabular}{ |p{.15\textwidth}|*{2}{P{.168\textwidth}|}*{2}{P{.099\textwidth}|}P{.168\textwidth}|}
    \hline
     & \textbf{Asus Tinker Edge R} & \textbf{Google Coral Dev Board} & \multicolumn{2}{c|}{\textbf{Nvidia Jetson Nano}} & \textbf{Arduino Nano 33} \\
    \hline
    \acs{API} & RKNN Toolkit & PyCoral/ \acs{TF} Lite & Tensor-RT & TF-TRT & \acs{TF} Lite micro \\
    \hline
    \acs{TF} 1 (frozen graph) & \faThumbsOUp & \faThumbsDown & \faThumbsOUp & \faThumbsOUp & - \\
    \acs{TF} 1 (saved model) & \faThumbsDown & \faThumbsDown & \faThumbsOUp & \faThumbsOUp & - \\
    \acs{TF} 2 (saved model) &\faThumbsDown & \faThumbsDown & \faThumbsOUp & \faThumbsOUp & - \\
    \acs{TF} Lite & \faThumbsOUp & \faThumbsDown  & \faThumbsOUp & \faThumbsDown & - \\
    \acs{TF} Lite quant. (8-bit quantized model) & \faThumbsOUp & \faThumbsOUp & \faThumbsDown & \faThumbsDown & \faThumbsOUp \\
    \hline
    \end{tabular}
    \label{tab:optcomp}
\end{table*}

\subsubsection{Tensorflow}
TensorFlow~\cite{tensorflow2015-whitepaper} is an open-source software library for numerical computation using data flow graphs. Nodes in the graph represent mathematical operations, while the graph edges represent the multidimensional data arrays (tensors) that flow between them. This flexible architecture deploys computation to one or more CPUs or GPUs without rewriting code. Tensorflow is the main framework we used for exporting pretrained  models for inferencing on the edge devices. As a framework, it is already optimized for systems with Nvidia graphics cards in combination with their neural network library \acs{CUDNN}. Nvidia's Jetson Nano is therefore capable of running networks on its integrated \ac{GPU}, giving it an advantage over the other devices that perform inference solely on the \ac{CPU}. 

\subsubsection{TensorRT}
NVIDIA TensorRT~\cite{tensorrt} is a C++ library that facilitates high performance inferencing on NVIDIA graphics processing units (GPUs). TensorRT takes a trained network, which consists of a network definition and a set of trained parameters, and produces a highly optimized runtime engine which performs inference for that network.
TensorRT applies graph optimizations, layer fusion, among other optimizations, while also finding the fastest implementation of that model leveraging a diverse collection of highly optimized kernels. 
TensorRT supports parsing models in ONNX, Caffe and UFF format.
To convert a model from Tensorflow or other frameworks, external tools like the \textit{tf2onnx} command line tool or \textit{uff} library have to be used~\cite{OnnxTensorflowonnx2021}.

Additionally, models can be optimized with TensorFlow with TensorRT (TF-TRT) optimization, which is included in the Tensorflow Python \ac{API}.

\subsubsection{Tensorflow Lite}
TensorFlow Lite was developed to run machine learning models on microcontrollers and other \ac{IoT} devices with only few kilobytes of memory~\cite{TensorFlowLiteMicrocontrollers}. Hence, runtime environments are offered for various platforms covering iOS, Android, embedded Linux and microcontrollers with \ac{API} support for Java, Swift, Objective-C, C++, and Python~\cite{TensorFlowLiteGuide}. TensorFlow Lite uses FlatBuffers as the data serialization format for network models, eschewing the Protocol Buffers format used by standard TensorFlow models. After building a model in the main Tensorflow framework, it can be converted to TFLite format with the inbuilt converter, supposing all operations used by the model are supported.
Unsupported operations can be manually implemented~\cite{TensorFlowLiteConverter}.

During the conversion process, it is also possible to quantize the model, which is needed for deploying to micro-controllers or devices utilizing the Google Edge TPU. To do that a Tensorflow Lite model must be ''compiled'' to be compatible with the Google Edge TPU, e.g. with the Edge TPU Compiler command line tool by Coral~\cite{EdgeTPUCompiler}.
In addition to being fully 8-bit quantized, only supported operations must be included in the model and the Tensor sizes must be constant at compile-time while only being 1, 2 or 3-dimensional (further dimensions are allowed as long as only the three innermost dimensions have a size greater than one)~\cite{TensorFlowModelsEdge}.
\\
For the model to be used in the C++ \ac{API}, which is required by the Arudino Nano 33 BLE, the TFLite model must be converted to a C array. This is done by using a hexdump tool, e.g. using the ''xxd -i'' Unix command.

\subsubsection{RKNN-Toolkit}
Rockchip offers its RKNN-Toolkit framework for running optimized inference on their \ac{AI} accelerated processors~\cite{RockchipQuickStart}.
It allows conversion from various frameworks (as seen in Table~\ref{tab:devspec}) to its own ''.rknn'' format.
The resulting models can then be run using the same framework.

\subsection{Model formats}
Pretrained Tensorflow models can come in a variety of different formats~\cite{ModelFormatsTensorFlow}.
The main ones used for deploying are \textit{SavedModel}, \textit{TF1 Hub} format, \textit{Frozen Graph} and \textit{TFLite} format~\cite{ModelFormatsTensorFlow}.
Additionally, TF-TRT and RKNN-Toolkit use other formats to run an accelerated inference on the dedicated unit.

\subsubsection{SavedModel Format} 
It is the recommended format for exporting models~\cite{SavedModelsTFHub}. It contains information on the complete Tensorflow program, including trained parameter values and operations.
Loading and saving a model can be done via one command without further knowledge about its internal structure. By default, a saved model cannot be trained any further after saving it. 
\subsubsection{TF1 Hub Format} 
The TF1 Hub format is a custom serialization format used in by TF Hub library~\cite{SavedModelsTFHub}. The TF1 Hub format is similar to the \textit{SavedModel} format of TensorFlow 1 on a syntactic level (same file names and protocol messages) but semantically different to allow for module reuse, composition and re-training (e.g., different storage of resource initializers, different tagging conventions for metagraphs).
To support loading a model in this format in the \ac{TF} 2 Python \ac{API}, an additional module has to be imported~\cite{TF1HubFormat, SavedModelsTFHub}.
\subsubsection{Frozen Graph Format} 
Frozen Graph format is the deprecated format. In contrary to the \textit{TF1 Hub} format, it saves the parameters as constants, bound to their respective operations. A \textit{Frozen Graph} can't be trained after it's loaded, though converting it to \textit{TF1 Hub} format is possible (yet still not trainable)~\cite{TfCompatV1}.

\subsubsection{TFLite Format} 
TFLite (".tflite") is used to deploy models and make them available on devices with limited resources~\cite{TensorFlowLiteGuide}. A model must hence be converted from a \ac{TF} model with one of the previous formats. Before converting, it must be considered that the TFLite runtime platform for performing the inference has only limited support for Tensorflow operations. Unsupported operations can therefore be explicitly implemented as custom operations in the runtime framework.
During conversion, the model can be quantized to reduce model size and inference speed.
\\

Table~\ref{tab:optcomp} provides a summary of the optimization compatibility for various pre-trained Tensorflow models on different devices. By looking at it, we can conclude that for a comparison using every device's full potential, a model must be available in Tensorflow Lite (and 8-bit quantized) and one of the other Tensorflow formats.
The Raspberry Pi is not included in this overview as it doesn't have a dedicated unit for which a model can be optimized for.
As we're focusing on IoT devices, we want to pick models that are commonly used in this category i.e. for image classification. The Tensorflow 1 Detection Zoo and the Tensorflow Model Garden therefore provide a variety of image classification models with different sizes and formats~\cite{TensorflowModels2021}. 


\section{Methodology}
\label{sec:methodology}
\begin{figure}[h!]
\centering
\resizebox{\columnwidth}{!}{%
\begin{tikzpicture}[scale=1/2]
    \node(dev)[draw, text depth=1.4cm, minimum width=5.6cm] at (0,3){\ac{SoC} edge device};
    \node(cdb)[draw, rectangle, anchor=south east, fill=black!20, minimum width=2.6cm] at ([xshift=-0.15cm,yshift=+0.25cm]dev.south){Coral Dev Board};
    \node(jn)[draw, rectangle, anchor=south west, fill=black!20, minimum width=2.6cm] at ([xshift=+0.15cm,yshift=+0.25cm]dev.south){Jetson Nano};
    \node(ter)[draw, rectangle, anchor=south east, fill=black!20, minimum width=2.6cm] at ([xshift=-0.15cm,yshift=+1.5cm]dev.south){Tinker Edge R};
    \node(rpi)[draw, rectangle, anchor=south west, fill=black!20, minimum width=2.6cm] at ([xshift=+0.15cm,yshift=+1.5cm]dev.south){Raspberry Pi};
    \node(sftpcli)[draw, rectangle, anchor=south west] at (dev.north west){SFTP Client};
    \node[draw, rectangle, anchor=south west] at (sftpcli.north west){NTP Client};
    \node[draw, rectangle, anchor=south east, fill=blue!20] at (dev.north east){\textit{test.py}};
    \node(pow)[draw, rectangle, minimum width=3cm] at (10,0){HM310P Power Supply};
    \node(mon)[draw, minimum width=4.75cm, fill=black!20] at (20,1.5){Monitoring device};
    \node(sftpsrv)[draw, rectangle, anchor=south west] at (mon.north west){SFTP Server};
    \node[draw, rectangle, anchor=south west] at (sftpsrv.north west){NTP Server};
    \node[draw, rectangle, anchor=south east, fill=blue!20] at (mon.north east){\textit{serial\_reader.py}};
    \draw[<-, thick] (dev.south) -- (0,0) |- node[below,midway]{DC Power}(pow.west);
    \draw[-, thick] ([yshift=+1.5cm]dev.east) -- node[above,midway,text width=6cm,align=center]{\acs{NTP} sync., dataset file access over \acs{SFTP} (over Ethernet)}(20,4.5) -| (mon.north);
    \draw[->, thick] (pow.east) -- (20,0) -| node[below,text width=4cm,align=center]{Power measurement data (over Serial USB)}(mon.south);
\end{tikzpicture}%
}

\centering
\caption{High-level overview of the testing setup showcasing the workflow between each of its components.}
\label{fig:setup}
\end{figure}
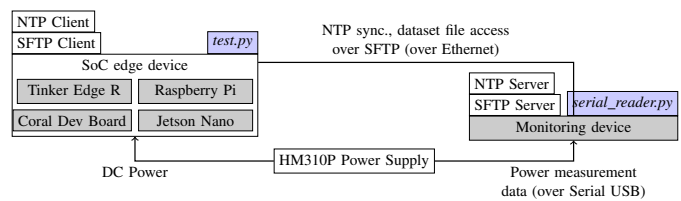
\noindent

For measuring time, power consumption and accuracy, we developed a python framework containing two main scripts: i) \textit{test.py} for running the models on the devices and getting inference data, and ii) \textit{serial\_reader.py} for getting power measurement data to be run on the monitoring device (a system used for collecting monitoring data from all the edge devices) as depicted in Figure~\ref{fig:setup}.

Power measurement and inference are performed on separate devices to avoid consumption of computing resources on the test device, therefore the clocks are coordinated. To accomplish this, we use the systemd\cite{Systemd} implementation of the \acl{NTP}, which is able to achieve better than 1ms deviation in local networks~\cite{NetworkTimeProtocol}.
The monitoring device therefore runs a \ac{NTP} server, for which the edge device is configured to request the current time from. For higher accuracy, the devices are directly connected via Ethernet as shown in Figure~\ref{fig:setup}.

In the following subsections we describe the three important tasks conducted by the two developed python modules in more details.
\subsection{Inference time calculation using test.py}
The \textit{test.py} script runs on the edge device under test, evaluates a model in a chosen framework and generates timestamps during the process. The \textit{test.py} script allows inference of any platform-compatible model trained on the ImageNet~\cite{ILSVRC15} or COCO~\cite{lin2014microsoft} dataset. On some of the devices, the ImageNet dataset occupies more memory than available. The directory containing the images are therefore saved on the monitoring device and accessed over \ac{SFTP}. Inference platforms include Tensorflow, Tensorflow Lite, RKNN-tookit, TensorRT with ONNX so the script can be run on every device up for testing.  The resulting output tensors are saved into a list for each image batch. Model accuracy is calculated after the last inference is performed. The program also supports defining the batch size, input datatype, image dimensions and number of inferences that should be performed. Besides measuring the model performance for each device, which is the time a device took for each inference, we also generate inference-timestamps indicating the current workload of the device (i.e. when an inference starts and ends). This is necessary for knowing exactly what process we are monitoring while calculating performance and watts per time later on.

Labeled timestamps (e.g. \textit{inf\_start\_batch\_i} denoting the start time and \textit{inf\_end\_batch\_i} denoting the end time) are generated just before and after a loop in which the inference function is called a given amount of times.
Thus, the measured time only includes the inference performed by the framework, excluding image pre-processing and evaluation of the results.

Overall inference time $t_{\text{inf}}$, average inference time $\overline{t}_{\text{inf}}$ and overall time spent on the testing program for $N$ images $t_{\text{test}}$ can then be calculated from the generated file as following:
\begin{equation}
\begin{aligned}
    t_{\text{inf}} &= \sum_{i=1}^{N}{t_{\text{inf}\_\text{end}\_\text{batch}\_i} - t_{\text{inf}\_\text{start}\_\text{batch}\_i}} \\
    \overline{t}_{\text{inf}} &= \frac{t_{\text{inf}}}{N} \\
    t_{\text{test}} &= t_{\text{test}\_\text{end}} - t_{\text{test}\_\text{start}}
\end{aligned}
\end{equation}

\subsection{Power consumption calculation using serial\_reader.py}
\begin{figure}[t!]
    \centering
     \includegraphics[width=0.94\linewidth]{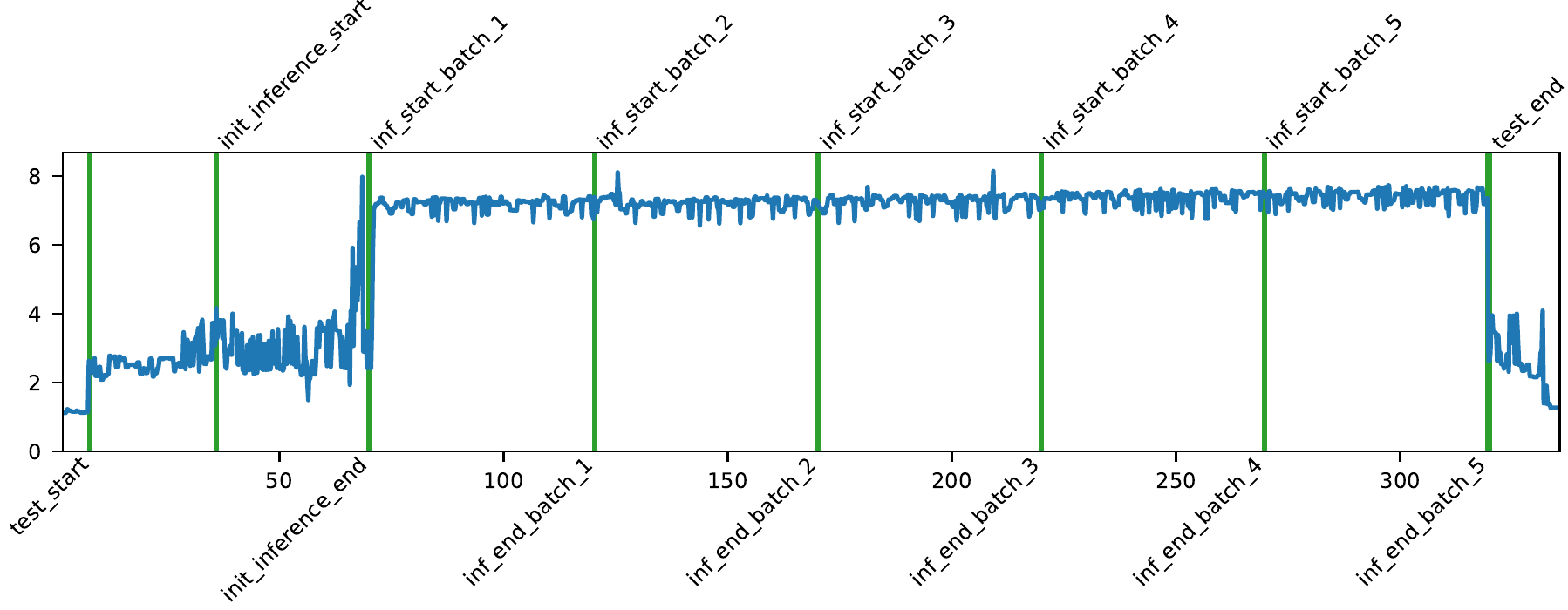}
    \caption{Power consumption (watts) over time (seconds) and labels during 1000 x 5 images inference on the Nvidia Jetson Nano with MobilenetV2 in Tensorflow}
    \label{fig:powexample}
\end{figure}
During the inference, the \textit{serial\_reader.py} script runs on a monitoring device to collect power consumption data. To measure power consumption, we calculate how much energy is consumed during the idle state, inference time and during the entire testing program. Knowing the exact value at any point in time during the process is mandatory to get a meaningful result. Therefore measuring with a wattmeter and reading the results by hand won't suffice. Instead, we measure voltage and current via a laboratory power supply with an integrated high precision power meter, the HM310P by Hammatek.
It provides a serial Modbus USB interface over which the registers containing configuration and status of the power supply can be read and set.
While no official software and documentation are available until now, a user-contributed documentation\cite{MckenmHanmaTekPSUCmd} shows an overview of the modbus accessible registers in the HM310P.
This made it possible to implement a python library \textit{hm310p.py} in the testing program, providing functions to control the device.

In a given interval, the \textit{serial\_reader.py} monitors the power consumption by requesting power, voltage and current from a given path to the HM310P device. Optionally, the script creates a live plot of the incoming data and/or saves the data with respective timestamps to a given path.
Afterwards, the generated inference-timestamps by the \textit{test.py} can be summarized and plotted with the same script as shown in Figure~\ref{fig:powexample} for Nvidia Jetson Nano when performing inference on 1000 x 5 images with MobilnetV2 in Tensorflow.

Overall energy consumption $wm_{\textbf{inf}}$ (in wattminutes) and average power $\overline{w}_{inf}$ (in watts) during inference are evaluated for an inference-timestamps file and power-timestamps file containing a list of tuples $T$. Each tuple contains time in seconds, current in amps, voltage in volts and power in watts.
Firstly, a list of power data can be extracted for each batch $i$, for which $W_i = \{w | (t,a,v,w) \in T, t_{\textbf{inf}\_\text{start}\_\text{batch}\_i} \leq t \leq t_{\text{inf}\_\text{end}\_\text{batch}\_i}\}$.
The respective values can then be calculated with
\begin{equation}
\begin{aligned}
    \overline{w}_i &= \frac{\sum\nolimits_{w \in W_i}{w}}{|W_i|} \\
    \overline{w}_{inf} &= \frac{\sum_{i=1}^{N}{\overline{w}_i}}{N} \\
    wm_{\text{inf}} &= \overline{w}_{\text{inf}} \times \frac{t_{\text{inf}}}{60}.
\end{aligned}
\end{equation}

\subsection{Model accuracy calculation using test.py}
The accuracy of models can change after being converted and optimized to work on a specific platform. Therefore it is important to include it into the comparison. For common tasks like image classification and object detection, there are metrics to measure how good an algorithm performs for a given dataset.

We use two metrics to indicate the accuracy of a model, Top-1 and Top-5 accuracy. For Top-1 accuracy, only the most probable class is compared with ground truth, whereas for Top-5 accuracy, the correct class just has to be among the top five guesses. We then consider the relation of correctly detected images to all images using the following equation:
\begin{equation}
    \begin{aligned}
        \text{Top} = \frac{\text{correct results}}{\text{all results}}
    \end{aligned}
\end{equation}

\noindent
During evaluation in the \textit{test.py} script, we calculate this value by labeling the resulting tensor values with their indices and then sorting them by weight as shown in Figure~\ref{fig:intimgdet}.
\begin{figure}[t!]
\resizebox{\columnwidth}{!}{%
\begin{tikzpicture}[scale=1/2]
    \node[draw] at (1,4.5){0.000};
    \node[draw] at (1,3.5){0.000};
    \node[draw] at (1,2.5){0.000};
    \node[draw] at (1,1.5){0.000};
    \node[draw](a0) at (1,0.5){0.000};
    \node[anchor=west,align=left] at (2,4.5){0: tench, Tinca tinca};
    \node[anchor=west,align=left] at (2,3.5){1: goldfish, Carassius auratus};
    \node[anchor=west,align=left] at (2,2.5){2: great white shark, \dots};
    \node[anchor=west,align=left] at (2,1.5){3: tiger shark, Galeocerdo cuvieri};
    \node[anchor=west,align=left](a1) at (2,0.5){4: hammerhead, hammerhead shark};
    \node[anchor=north] at (a0){\textbf{\vdots}};
    \node[anchor=north] at (a1){\textbf{\vdots}};
    \node[draw] at (16,4.5){0.539};
    \node[draw] at (16,3.5){0.085};
    \node[draw] at (16,2.5){0.081};
    \node[draw] at (16,1.5){0.068};
    \node[draw](a2) at (16,0.5){0.011};
    \node[anchor=west,align=left,draw=blue] at (17,4.5){66: sea snake};
    \node[anchor=west,align=left] at (17,3.5){55: hognose snake, puff adder, sand viper};
    \node[anchor=west,align=left] at (17,2.5){63: rock python, rock snake, Python sebae};
    \node[anchor=west,align=left] at (17,1.5){59: water snake};
    \node[anchor=west,align=left](a3) at (17,0.5){68: diamondback, diamondback rattlesnake, \dots};
    \node[anchor=north] at (a2){\textbf{\vdots}};
    \node[anchor=north] at (a3){\textbf{\vdots}};
    \draw[->,line width=2pt,fill=] (12,2.5) -- (14,2.5);
\end{tikzpicture}%
}
\caption{Interpreting the result of an image detection model}
\label{fig:intimgdet}
\end{figure}
If the ground truth value for the image equals the first value or is among the first five values, the amount of \textit{correct results} is incremented respectively for Top-1 and Top-5 accuracy. With a significant amount of samples, we get an approximation for the accuracy of a model.
The ImageNet dataset ILSVRC2012~\cite{ILSVRC15} therefore provides 50,000 images for validation with ground-truth data, from which we're going to use 5,000 samples for each run.
\\
\\
The described methodology allows testing any device supporting the respective Python libraries of the \textit{test.py} script.
This is not the case for the Arudino Nano 33 BLE, so power consumption and inference time plus accuracy is measured separately.
For measuring inference time, a C++ script is provided to be run on the arduino, implementing the TFLite C++ library. The script sets up the MobileNetV1 network and waits for the input image to be received over serial USB, which are then put into the placeholders of the input tensor. Inference time is measured on the arduino and sent back over serial USB afterwards, together with the output tensor after the inference is complete. Transmission and evaluation of the dataset data is handled by the \textit{test.py} script, which is run on the monitoring device this time.

\section{Experimental Configurations}
\label{sec:exp_config}
The previously described methodology is used with different configurations (model, edge device, framework, given amount of images from the ImageNet data) combinations for evaluations. For more precise results, one test run with 5,000 images for a decent model accuracy approximation and one with 5 images repetitively inferenced 1,000 times to further counteract inaccuracies due to possible delays when reading power measurement data from the power supply are conducted.

During all measurements, unnecessary processes like graphical user interfaces as well as internal devices like wireless \ac{LAN} modules are disabled to minimize power consumption and processing power.

It may also be possible that an inference, performed solely on the \ac{CPU} may be more energy-efficient than with using the dedicated unit. We therefore also test the devices using the ''native'' Tensorflow and Tensorflow Lite framework without optimizations for a dedicated unit.
\\
In the results section, the main focus will be on the following collected data:
\begin{itemize}
    \item Average time spent on inference of one of 5,000 images
    \item Total time spent on inference of 5,000 images
    \item Power during idle state (LAN on and off)
    \item Average power during inference of 1,000 x 5 images
    \item Total power consumption during inference of 1,000 x 5 images
    \item Accuracy for each platform-device combination
\end{itemize}

\subsection{Models for evaluation}
We pick models that can be optimized for as many devices as possible. For an extensive comparison, four model configurations are considered.
The most limiting factor for finding a common model is for the network to fit on the Arudino Nano 33 BLE with only 1MB of \ac{RAM}.
The only image classification network from the Tensorflow Model Garden that is within that range is the 8-bit quantized "MobileNet\_v1\_0.25", available in \textit{TFLite} and \textit{TF1 Hub} format (\textbf{MobileNetV1 Quant. Lite})~\cite{TensorFlowModelGarden2021}.

\textit{Ergo}, the quantized TFLite model can also be deployed to the Asus Tinker Edge R, the Google Coral Dev Board and the Raspberry Pi.
Though being capable of running on the Arudino Nano 33 BLE, its small size may lead to less accurate results on the other devices.
An inference for one image takes less than 2ms on the Google Coral Dev Board. With an offset of up to 1ms between power measurement timestamps and inference timestamps, up to 50\% of the collected power consumption data may not be assignable to the current workload and therefore be invalid.

Selecting a bigger size model that takes longer for the evaluation does increases the accuracy of measuring inference time, since the relation of the actual inference to unintentionally included function calls (like time()) increases. Hence, we also use a larger MobileNetV2 model with 12x as many parameters, also available on the Tensorflow Model Garden.

Ascribed to the incompatibility of non-quantized models on the Google Coral Dev Board and quantized models on the Nvidia Jetson Nano, there are two main test runs for the model. The first one includes the "float\_v2\_1.4\_224" Frozen Graph to be evaluated on the Asus Tinker Edge R, the Raspberry Pi and the Nvidia Jetson Nano (\textbf{MobileNetV2}). The second, includes the model's corresponding 8-bit quantized version to be run on the Google Coral Dev Board instead of the Nvidia Jetson Nano (\textbf{MobileNetV2 Quant. Lite}).
\begin{figure*}[t]
     \centering
     \includegraphics[width=1\linewidth]{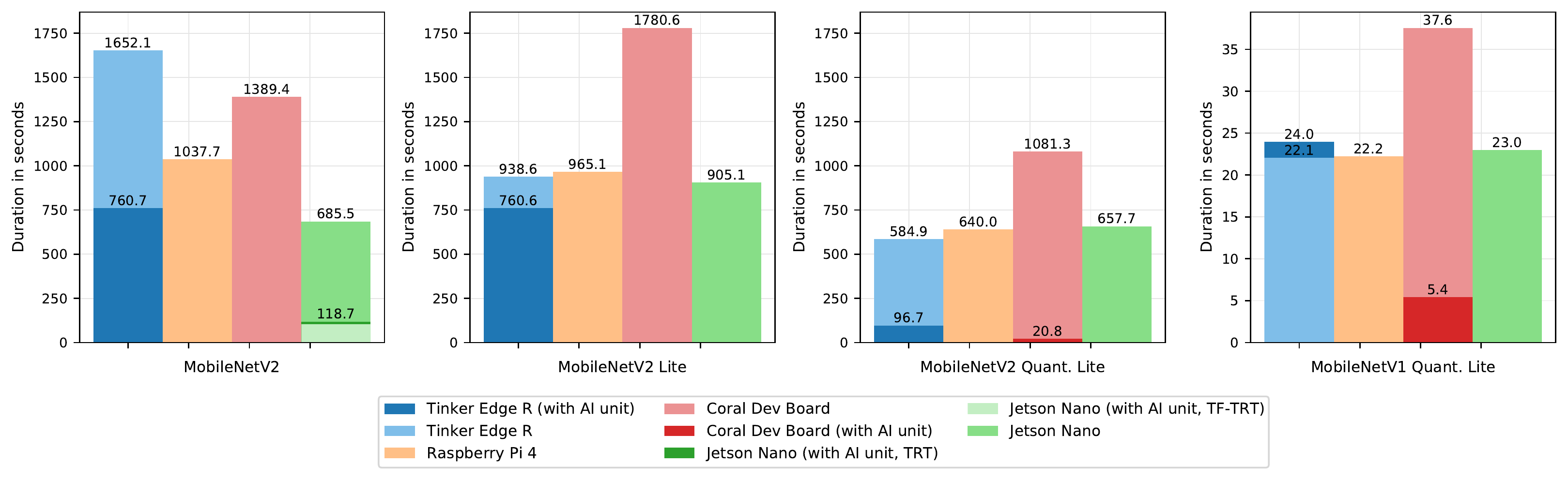}
    \caption{Time taken for performing inference (1 $\times$ 5000 images test run) on various devices when using only \ac{CPU}, as well as when utilizing the dedicated \ac{AI} unit across different models and framework. }
    \label{fig:time}
\end{figure*}

We also consider the non-quantized version to see the impact of quantization and use of TFLite (\textbf{MobileNetV2 Lite}).

All previously discussed image classification models are trained on the ImageNet ILSVRC2012 dataset~\cite{ILSVRC15}.

\section{Results}
\label{sec:results}
After performing the experiments, we get an insight into the strengths and weaknesses of each device. In the following subsections, we discuss how the devices compare in respect to the evaluation scenarios.

\subsection{Inference performance}
Figure~\ref{fig:time} shows the time taken for performing inference on various devices when using only \ac{CPU}, as well as when utilizing the dedicated \ac{AI} unit across different models.

First, we will take a look at the inference performance without utilization of the dedicated \ac{AI} units. Although this information may be irrelevant for actual use cases with those devices, it may be useful when comparing other devices utilizing the same \acp{CPU} that do not include an \ac{AI} unit. Additionally, it gives a hint on the performance difference for Tensorflow and Tensorflow Lite framework and by how much the performance increases when using the respective dedicated units.

\subsubsection{Inference performance without utilization of the dedicated \ac{AI} units}

For \textit{MobileNetV2},  Nvidia Jetson Nano outperforms its competition in terms of inference performance with \texttt{685.490} seconds (\texttt{$\sim$0.137} seconds per inference) followed by Rasphberry Pi 4 with \texttt{1037.7} seconds (\texttt{$\sim$0.21} seconds per inference), then Coral Dev board with \texttt{1389.4} seconds (\texttt{$\sim$0.28} seconds per inference), and lastly Asus Tinker Edge R  with \texttt{1652.1} seconds (\texttt{$\sim$0.33} seconds per inference). It shows that, Nvidia Jetson Nano is almost \textbf{2.5x} faster than the Asus Tinker Edge R. 

Besides determining the best performing devices, running the same model in Tensorflow and Tensorflow Lite Runtime seemingly makes a significant difference (\textit{MobileNetV2} vs \textit{MobileNetV2 Lite}). From the Figure~\ref{fig:time} we can see that, although in case of \textit{MobileNetV2 Lite}  Nvidia Jetson Nano performs best with \texttt{905.1} seconds (\texttt{$\sim$0.181} seconds per inference) followed by Asus Tinker Edge R with \texttt{938.6} seconds (\texttt{$\sim$0.187} seconds per inference), then Raspberry Pi 4  with \texttt{965.1} seconds (\texttt{$\sim$0.193} seconds per inference), and lastly Coral Dev board  with \texttt{1780.6} seconds (\texttt{$\sim$0.356} seconds per inference), but when compared with \textit{MobileNetV2} model, The Tinker Edge R (\texttt{43\%} reduction in time) and Raspberry Pi (\texttt{7\%} reduction in time) perform better with Tensorflow Lite, while the Coral Dev Board (\texttt{32\%} increase in time) and Jetson Nano (\texttt{28\%} increase in time) perform worse.

In case of the quantized TFLite models (\textit{MobileNetV2 Quant. Lite} and \textit{MobileNetV1 Quant. Lite}), the Asus Tinker Edge R has the best performance (\texttt{584.9} seconds for \textit{MobileNetV2 Quant. Lite}  and \texttt{22.2} seconds for \textit{MobileNetV1 Quant. Lite}), although performing the inference for MobileNetV1 on its \ac{NPU} (\texttt{24.0} seconds) is actually slower than on its \ac{CPU}. The most probable reason for this behavior is the small size of the model, considering that the Tinker Edge R performed significantly better for the other models when utilizing its \ac{NPU}.
It is followed by Rasphberry Pi 4 with \texttt{640.0} seconds for \textit{MobileNetV2 Quant. Lite} and \texttt{22.2} seconds for \textit{MobileNetV1 Quant. Lite} which is the same as that of Asus Tinker Edge R, then Nvidia Jetson Nano with \texttt{657.7} seconds for \textit{MobileNetV2 Quant. Lite} and \texttt{23.0} seconds for \textit{MobileNetV1 Quant. Lite} and lastly, Coral Dev board with \texttt{1081.3} seconds for \textit{MobileNetV2 Quant. Lite} and \texttt{37.6} seconds for \textit{MobileNetV1 Quant. Lite}. One can conclude that Asus Tinker Edge R is almost \textbf{1.8x} faster for \textit{MobileNetV2 Quant. Lite} and \textbf{1.7x} faster for \textit{MobileNetV1 Quant. Lite} than Coral Dev board. 

Additionally, when comparing \textit{MobileNetV2 Lite} with \textit{MobileNetV2 Quant. Lite}, shows that model quantization results in performance increment for all the devices. From the Figure~\ref{fig:time} we can see that, for inference, there is a \texttt{27.3\%} reduction in inference time for Jetson Nano, while \texttt{37.6\%} for Asus Tinker Edge R, \texttt{39.3\%} for Coral Dev board, and \texttt{33.7\%} for Raspberry Pi 4. Coral Dev board saw the most improvement in performance by model quantization.

Lastly, the Arduino lacks behind with its microcontroller processor, taking \texttt{57534.297} seconds to inference 5000 images (\texttt{$\sim$11.509}  seconds per inference). This may seem like an enormous drawback compared to the other \acp{SoC}, but is actually quite acceptable for actual use cases when also taking power consumption during inference and idle state into account.



\begin{figure*}[t]
     \centering
     \includegraphics[width=1\linewidth]{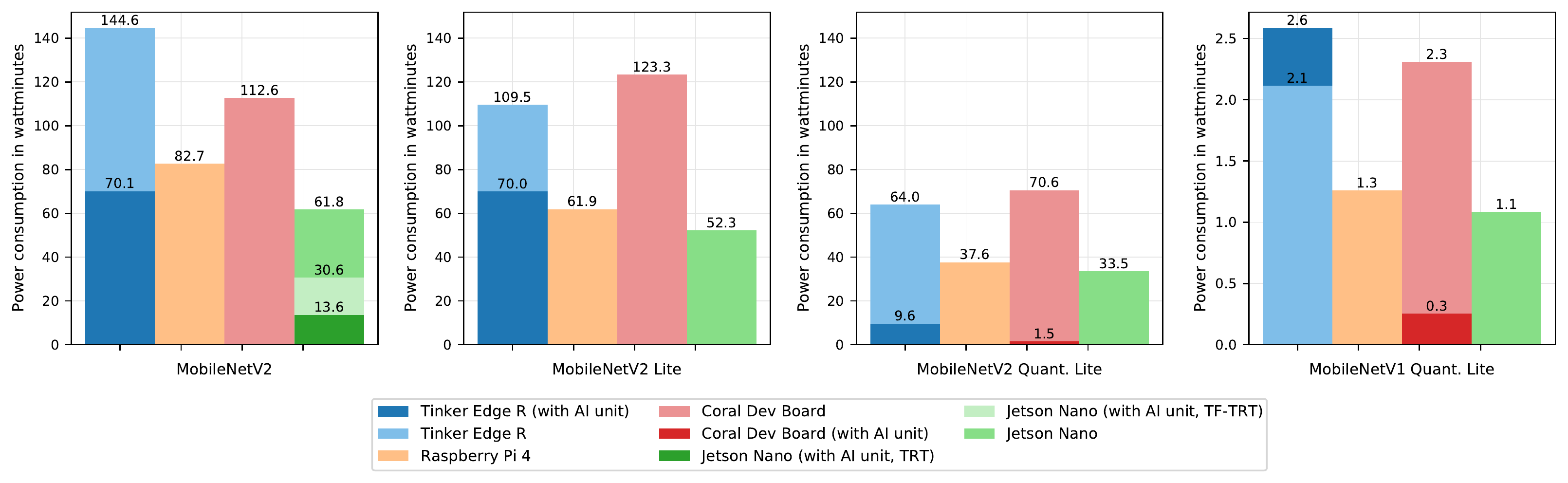}
    \caption{Power consumption (in watt minutes) when performing inference (1 $\times$ 5000 images test run) by various devices when using only \ac{CPU}, as well as when utilizing the dedicated \ac{AI} unit across different models and frameworks. }
    \label{fig:power_consumption}
\end{figure*}
\begin{table*}[h!]
    \centering
    \caption{Power consumption during idle state (in Watts) for various devices. }
    \label{table:power_idle_state}
    \begin{tabular}{ |p{.1\textwidth}|*{5}{P{.15\textwidth}|}}
    \hline
    & \textbf{Asus Tinker Edge R} & \textbf{Raspberry Pi 4} & \textbf{Coral Dev Board} & \textbf{Nvidia Jetson Nano} & \textbf{Arduino Nano 33}\\
    \hline
    \textbf{Idle (LAN)}    & 4.932 & 2.643 & 3.081 & 1.391 & \faClose  \\
    \hline
    \textbf{Idle (no LAN)} & 4.776 & 2.100 & 2.757 & 0.903 & 0.036 \\
    \hline
    \end{tabular}

\end{table*}
\subsubsection{Inference performance with utilization of the dedicated \ac{AI} units}

Based on the optimization compatibility of \ac{AI} units for various pre-trained Tensorflow models on different devices listed in the Table~\ref{tab:optcomp}, the results are shown in Figure~\ref{fig:time}. By looking at the performance, the respective devices naturally perform significantly better when optimized with their respective \ac{API} (except for Tinker Edge R with MobileNetV1).

The Google Coral Dev Board delivers the best performance for MobileNetV2 (using \textit{MobileNetV2 Quant. Lite}) with the inference time of \texttt{20.788} seconds (\texttt{$\sim$0.004} seconds per inference). Though it must be considered that the Nvidia Jetson Nano doesn't have the advantage of running the quantized version of the model. If that is taken into account, the Jetson Nano outperforms the other devices when it comes to inference the non-quantized version in Tensorflow (shown under \textit{MobileNetV2} in Figure~\ref{fig:time}). This is the case for both the Tensorflow-TensorRT optimized model with the inference time of \texttt{103.142} seconds (\texttt{$\sim$0.020} seconds per inference), as well as the TensorRT framework with the inference time of \texttt{118.737} seconds (\texttt{$\sim$0.023} seconds per inference), using the model in ONNX format.

We also see the conversion from Tensorflow to Tensorflow Lite (shown under \textit{MobileNetV2} and \textit{MobileNetV2 Lite} in Figure~\ref{fig:time}) keeps the structure of the model, since the MobileNetV2 model in Frozen Graph format ( taking \texttt{760.7} seconds) and TFLite format ( taking \texttt{760.6} seconds) perform similar to each other when ported to the RKNN-Toolkit for Asus Tinker Edge R.
This indicates that the previously mentioned performance deviations between the two frameworks are due to their implementation, rather than potential changes to the model during conversion.

\subsection{Power consumption}
Next, we illustrate how energy efficient the devices are when \textit{continuously} performing the computations. Figure~\ref{fig:power_consumption} shows the power consumption by various devices during inference when using only \ac{CPU}, and the dedicated \ac{AI} unit across different models. Additionally, for many use cases, deployed edge devices will not perform explicit computations continuously. Rather, a predefined number of maximum \ac{FPS} (or minute) might be anticipated to reduce energy use. Therefore it is worth looking at the power draw by the devices during their idle state i.e. while the device is not performing explicit computations. The Table~\ref{table:power_idle_state} shows the power consumption during idle state (in Watts) for various devices. 
In idle state, the Arduino Nano 33 BLE consumes the least energy of \texttt{0.036} Watts.  This is only around 0.04\% of the energy consumed by the most energy-efficient device evaluated here, the Jetson Nano (with LAN disabled).

\subsubsection{Power consumption without utilization of the dedicated \ac{AI} units}
From Figure~\ref{fig:power_consumption}, in overall, when using only CPU, the Jetson Nano consumes the least power for all models (\texttt{61.8} wattminutes for \textit{MobileNetV2}, \texttt{52.3} wattminutes for \textit{MobileNetV2 Lite}, \texttt{33.5} wattminutes for \textit{MobileNetV2 Quant. Lite} and \texttt{1.1} wattminutes for \textit{MobileNetV1 Quant. Lite}), followed by the Raspberry Pi (\texttt{82.7} wattminutes for \textit{MobileNetV2}, \texttt{61.9} wattminutes for \textit{MobileNet V2 Lite}, \texttt{37.6} wattminutes for \textit{MobileNetV2 Quant. Lite} and \texttt{1.3} wattminutes for \textit{MobileNetV1 Quant. Lite}) as shown in Figure~\ref{fig:power_consumption}. The trend we see here is very similar to the average power drawn by the devices in idle state (from Table~\ref{table:power_idle_state}).
However for the \textit{MobileNetV2 Lite} and  \textit{MobileNetV2 Quant. Lite} models, the Asus Tinker Edge R consumed less power (\texttt{109.5} wattminutes and \texttt{64.0} wattminutes respectively ) as compared to Coral Dev Board (\texttt{109.5} wattminutes and \texttt{64.0} wattminutes respectively) which is the opposite to their idle power consumption trend (Asus Tinker Edge R consuming more power than Coral Dev Board), as the Tinker Edge R's performance in Tensorflow Lite on the \ac{CPU} is sufficient to compensate for its higher power draw in idle state.

Lastly, the Arduino falls behind by consuming approximately \texttt{61.879} watthours of power since it needs a longer time for doing inference.

\subsubsection{Power consumption with utilization of the dedicated \ac{AI} units}
Analogously to inference performance, the devices are more energy efficient when using their \ac{AI} unit (except for the Tinker Edge R with \textit{MobileNetV1}). From Figure~\ref{fig:power_consumption}, the Coral Dev Board consumes the least power with just \texttt{1.543} wattminutes when considering across all \textit{MobileNetV2} models  
and \texttt{0.254} wattminutes for the \textit{MobileNetV1 Quant. Lite} model. 
Across all \textit{MobileNetV2} models, the Tinker Edge R is the second best option with \texttt{9.619} wattminutes (shown under \textit{MobileNetV2 Quant. Lite} in Figure~\ref{fig:power_consumption}), although the Jetson Nano is not far behind with \texttt{13.630} wattminutes (shown under \textit{MobileNetV2} in Figure~\ref{fig:power_consumption}). In this regard, the Jetson Nano once again has the lowest value when looking at the non-quantized version only. For the Jetson Nano, it is also observed that, although it is faster to perform inference with the TF-TRT optimized model, but it consumes less power using the TensorRT framework (\texttt{30.645} vs. \texttt{13.640} wattminutes, shown under \textit{MobileNetV2} in Figure~\ref{fig:power_consumption}).





\subsubsection{Power consumption prediction}
\begin{figure}[t]
     \centering
     \includegraphics[width=1\linewidth]{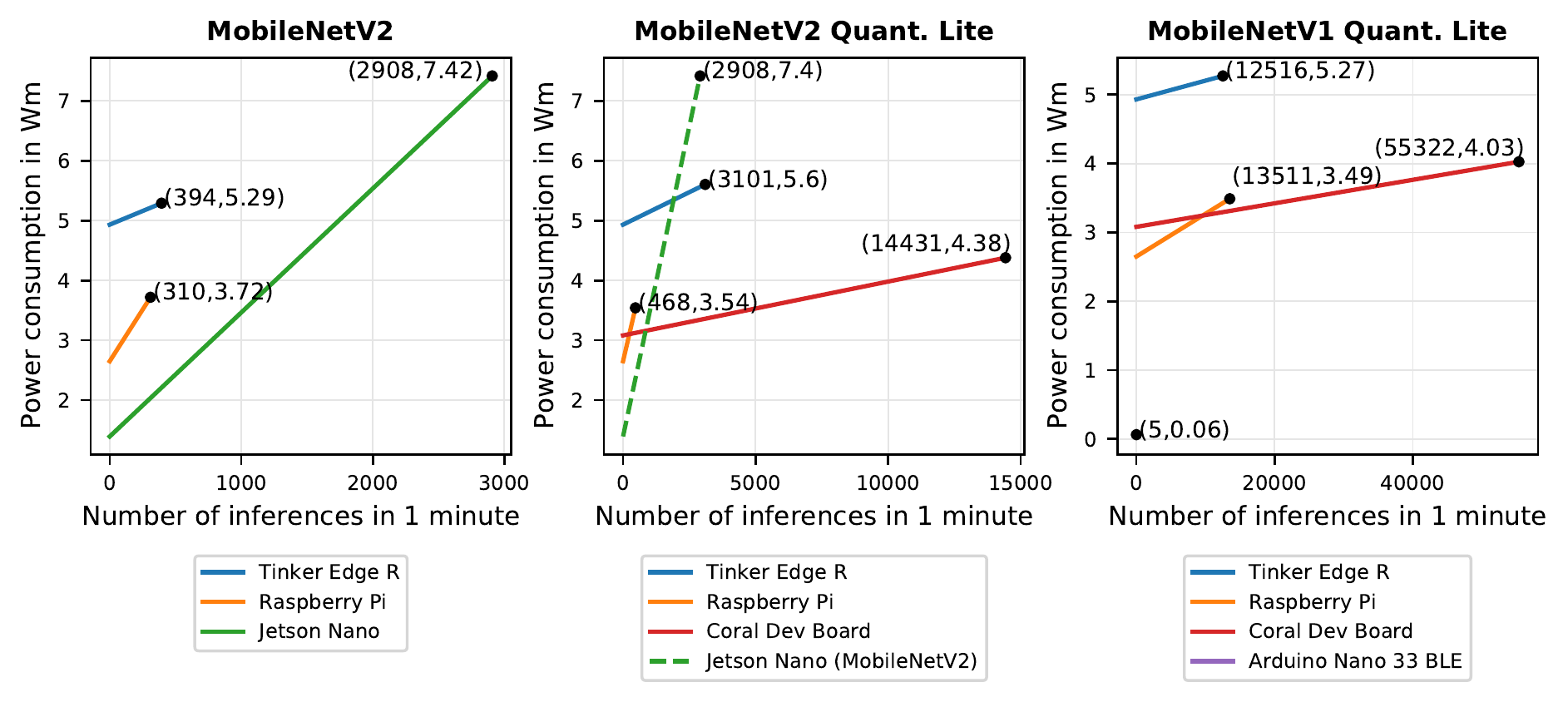}
    \caption{Sparsely predicted power consumption at different inference rates for different models by various devices.}
    \label{fig:powinfnum}
\end{figure}
\noindent
Based on the data on energy consumption during the inference, we can now broadly predict the devices power consumption at different inference rates for different models as shown in Figure~\ref{fig:powinfnum}. The indicated values in the graphs shows the the maximum number of inferences that can be performed in one minute and power consumption (in wattminutes) during those inferences by the respective devices.
Since power draw during initialization of the network, pre-processing and evaluation of the results are ignored, the actual power consumption is expected to be higher in a real-world scenario. For calculating the values, the best score in terms of power efficiency were taken for each model. As the \acp{SoC} had an Ethernet connection during testing, the respective values for idle power with \ac{LAN} are used.


The Raspberry Pi is observably more energy efficient than the Coral Dev Board when performing inference using the \textit{MobileNetV2 Quant. Lite} model (second sub-figure in Figure~\ref{fig:powinfnum}) at a inference rate of under \texttt{238} inferences per minute (\texttt{50.65}\% of the time spent on \ac{AI} computation). When also taking the Jetson Nano's results for the non-quantized Tensorflow version (\textit{MobileNetV2} in Figure~\ref{fig:powinfnum}) into account, it outperforms the other devices for an inference rate of under \texttt{853} inferences per minute (less than \texttt{29.3}\% of the time spent on \ac{AI} computation).

From all the results until now, the Arduino Nano 33 BLE did not hold up against the other devices. However, under different circumstances, the microcontroller performs significantly better i.e when inferences are performed in larger intervals. As seen from third sub-figure of Figure~\ref{fig:powinfnum}), at 5 frames per minute, the Arduino approximately consumes only \texttt{0.063} wattminutes which is considerably less than any other devices. This inference rate is also the maximum number of inferences that the device can perform in one minute for \textit{MobileNetV1}.

\subsection{Accuracy}
We now review the accuracy of the models, which may differ as a result of model conversion and optimization. There are no explicit details about the accuracy of the quantized versions of the models, though expected to be similar to the non-quantized versions~\cite{TensorFlowModelGarden2021}. Table~\ref{table:accuracy_models} shows the models accuracy's (top-1 and top-5) in different frameworks for different devices. 

\begin{table*}[h!]
    \centering
    \caption{Models accuracy's (top-1 and top-5) in different frameworks for different devices (1 x 5000 images).}
    \label{table:accuracy_models}
    \begin{tabular}{ |p{.17\textwidth}|*{4}{P{.085\textwidth}|}*{2}{P{.085\textwidth}|}P{.085\textwidth}|}
    \hline
    & \textbf{Claimed Accuracy} & \textbf{Asus} \newline \textbf{Tinker Edge R} & \textbf{Raspberry Pi 4} & \textbf{Google Coral Dev Board} & \multicolumn{2}{c|}{\textbf{Nvidia Jetson Nano}} & \textbf{Arduino} \newline \textbf{Nano 33 BLE} \\
    \hline
     & & \textbf{RKNN Toolkit} & \textbf{None/ TF} & \textbf{PyCoral/  TFLite} & \textbf{TF-TRT} & \textbf{TensorRT} & \textbf{TFLite Micro} \\
    \hline
    MobileNetV2             & 0.75, 0.925  & 0.734, 0.907 & 0.733, 0.908 & \faClose             & 0.733, 0.908 & 0.733, 0.908  & \faClose  \\
    \hline
    MobileNetV2 Lite        & 0.75, 0.925  & 0.734, 0.908 & 0.733, 0.908 & \faClose             & \faClose             & \faClose              & \faClose  \\
    \hline
    MobileNetV2 Quant. Lite & 0.75, 0.925  & 0.726, 0.904 & 0.727, 0.904 & 0.727, 0.904 & \faClose             & \faClose              & \faClose  \\
    \hline
    MobileNetV1 Quant. Lite & 0.395, 0.644 & 0.364, 0.613 & 0.361, 0.611 & 0.359, 0.612 & \faClose             & \faClose              & 0.361, 0.612 \\
    \hline
    \end{tabular}
    
\end{table*}
\noindent

Since only 5,000 out of 50,000 validation images were used, the claimed accuracy deviates from the test run in the native Tensorflow framework, which we can see in the Raspberry Pi's column of Table~\ref{table:accuracy_models}. No significant changes are observed when porting the models to the other frameworks, though results differ slightly when using RKNN-Toolkit, Tensorflow Lite Micro and when using the Edge TPU in Tensorflow Lite. More noticeable is the effect of quantization of the \textit{MobileNetV2} model. Due to less precise weights, accuracy decreases by about \texttt{0.6\%} for Top-1 and \texttt{0.4\%} for Top-5 accuracy.

\subsection{Performance on larger models}
Until now, the models used for testing were comparably small.
To demonstrate how model size affects the performance difference, the Raspberry Pi and Jetson Nano were additionally tested with an object detection model trained on the COCO~\cite{lin2014microsoft} dataset from the Tensorflow Detection Zoo. SSD MobilenetV2 COCO has more than 17M parameters, compared to 6.06M on MobilenetV2. 

Despite having no acceleration unit, the Raspberry Pi (taking \texttt{ 433.331} seconds for inference) now has similar performance to the Jetson Nano using the acceleration unit (taking \texttt{ 416.745} seconds for inference).

\begin{figure}[h!]
\begin{subfigure}{.49\linewidth}
    \includegraphics[width=\linewidth]{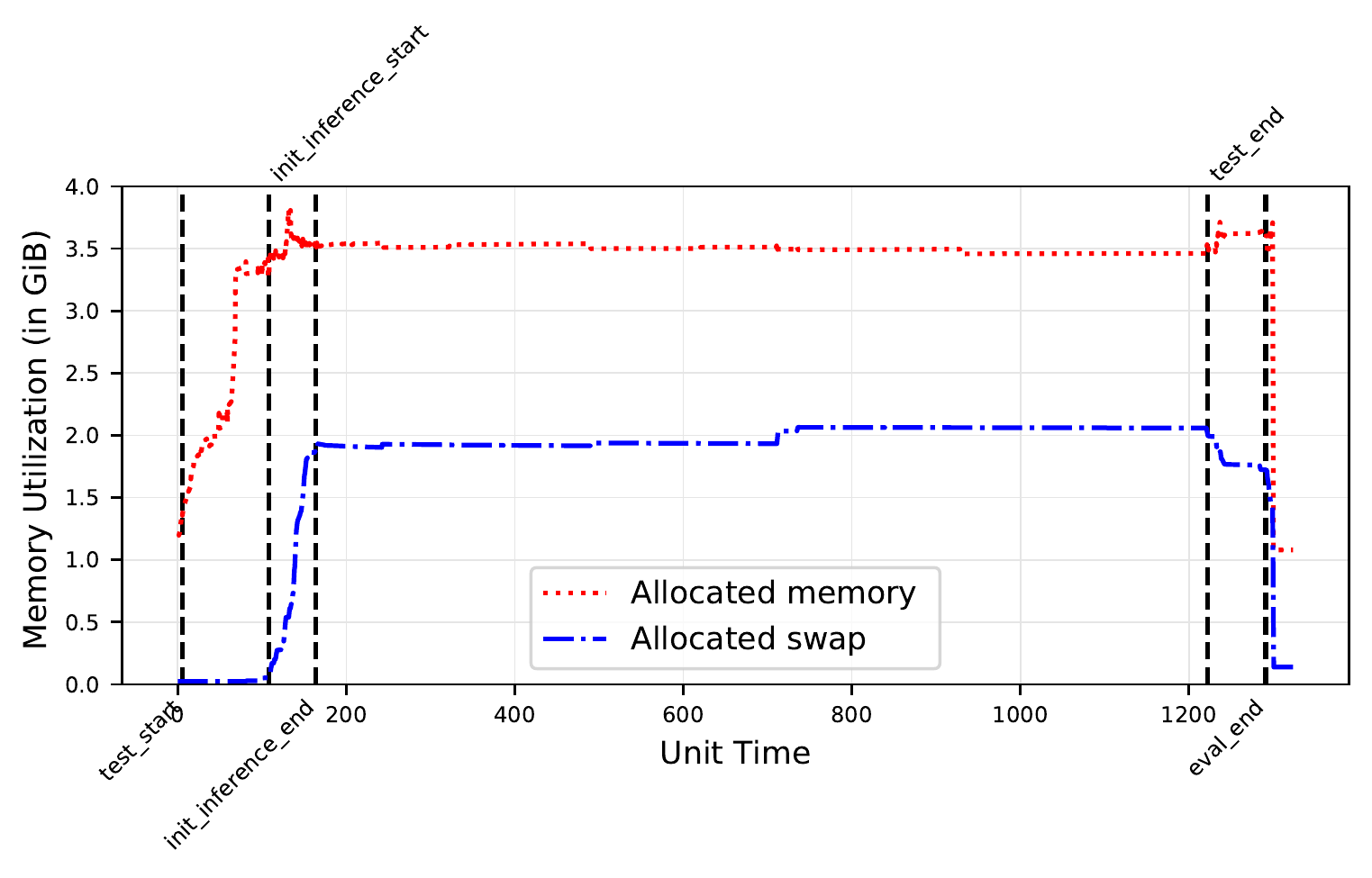}
    \caption{Jetson Nano}
    \label{fig:memdet_jetson}
\end{subfigure}
\begin{subfigure}{.49\linewidth}
    \includegraphics[width=\linewidth]{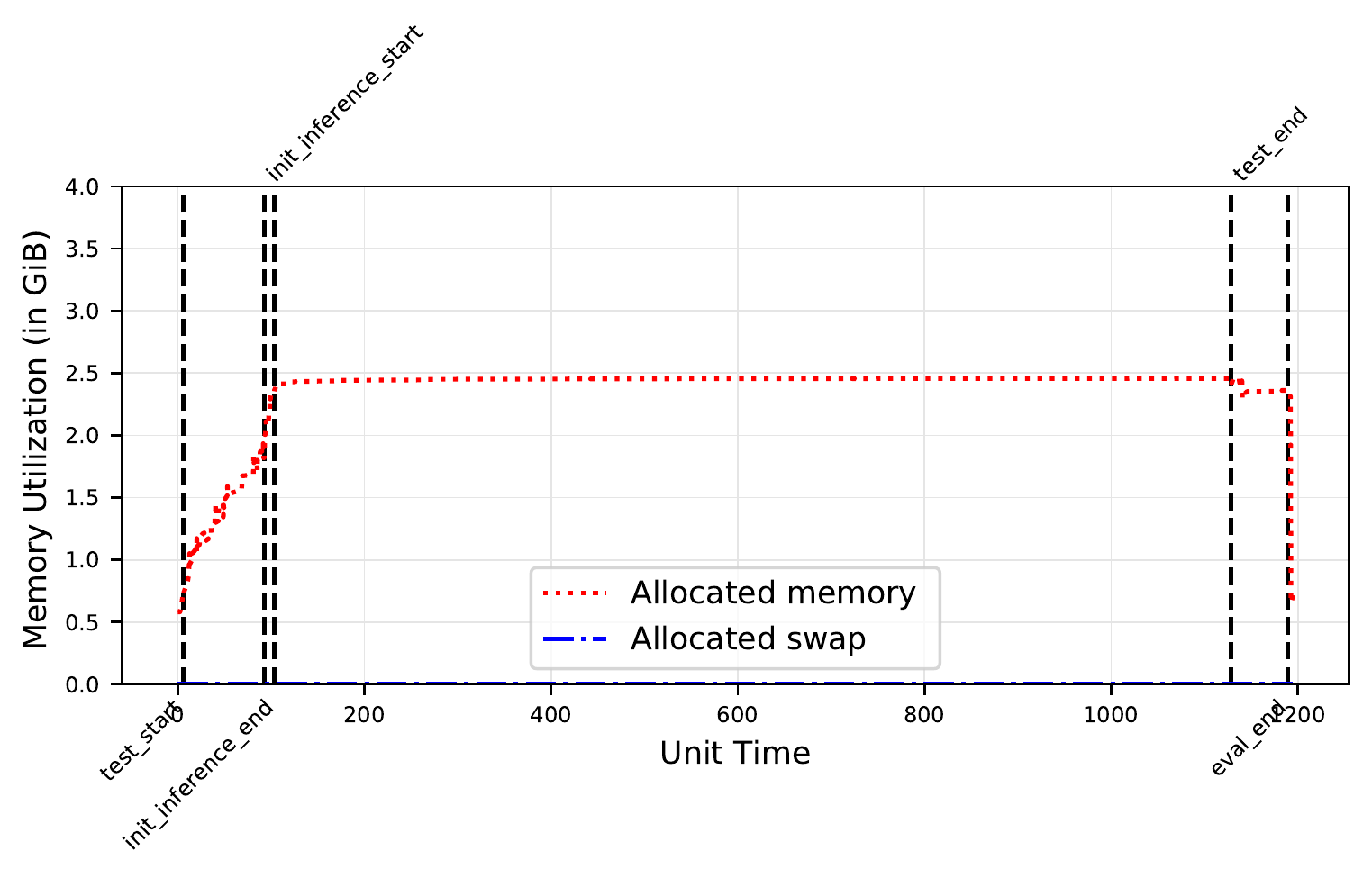}
    \caption{Raspberry Pi}
    \label{fig:memdet_rpi}
\end{subfigure}
\caption{Memory occupation (in GiB) over time (seconds) during 1 $\times$ 1000 images inference with SSD MobilenetV2 COCO in Tensorflow for two devices.}
\label{fig:memdet}
\end{figure}

This can be attributed to the fact that, memory on the Jetson Nano is shared between \ac{GPU} and \ac{CPU}. Figure~\ref{fig:memdet} shows the memory occupation over time for both the devices during 1 $\times$ 1000 images inference with SSD MobilenetV2 COCO in Tensorflow.  From Figure~\ref{fig:memdet_jetson}, we can observe that the internal \ac{RAM} for Jetson Nano does not suffice to load the model into memory during initialization inference. As a result, the swap space on the much slower SD card is additionally used, making an inference using the \ac{GPU} slower than when solely using the \ac{CPU} as is the case with Raspberry Pi (shown in Figure~\ref{fig:memdet_rpi}).

\section{Conclusion}
\label{sec:conclusion}
In this work,  we present and compare the performances in terms of inference time and power consumption of the four \acsp{SoC}:  Asus Tinker Edge R, Raspberry Pi 4, Google Coral Dev Board, Nvidia Jetson Nano,   and one microcontroller: Arduino Nano 33 BLE for different models and frameworks. We also provide a method for measuring power consumption, inference time and accuracy for the devices, which can be easily extended to other devices.

Noticeably, the results for each model turn out to be quite different, depending on model size (In terms of operations and parameters), quantization, framework and anticipated number of inferences per time.
Nevertheless we can draw some major conclusions for the following two main applications:
\begin{itemize}
    \item \textbf{Best performing device for continuous \ac{AI} computation}: The main factor here are the overall wattminutes consumed for inference a given number of images. For a Tensorflow model that can be quantized and converted to TFLite format, the Google Coral Dev Board delivers the best performance, both for inference time and power consumption. Not all models are developed to only include TFLite-compatible operations though. Therefore the Jetson Nano can run accelerated inference for Tensorflow models on its \ac{GPU}. And as the entire Tensorflow framework can utilize the \ac{GPU}, models can also be trained. While it might not be worth training a model on a single low-power Edge Device, distributed inference is already discussed by some recent papers~\cite{zhouDistributingDeepNeural}.
    
    \item\textbf{Best performing device for sporadic \ac{AI} computation}: When just sporadically performing \ac{AI} computations, power efficiency is mostly dependent on power draw during idle. For small enough models, the Arduino Nano 33 BLE is by far the most power efficient option, though respectively low inference rates might affect usability. On the \ac{SoC} side it depends on the model and the anticipated number of inferences per time. The Jetson Nano presumably outperforms the other devices for a low fraction of \ac{AI} computation time (less than \textit{29.3}\% of the time for MobileNetV2). Otherwise, the Google Coral Dev Board is the most power efficient.
\end{itemize}

Extending the work to include other \ac{SoC}s and evaluating using larger CNN models is prospective future work.

\section{Acknowledgement}
This work was supported by the funding of the German Federal Ministry of Education and Research (BMBF) in the scope of the Software Campus program.
\bibliographystyle{IEEEtran}  
\bibliography{bib}

\end{document}